\def\year{2018}\relax
\documentclass[letterpaper]{article} 
\usepackage{aaai18}  
\usepackage{times}  
\usepackage{helvet}  
\usepackage{courier}  
\usepackage{url}  
\usepackage{graphicx}  

\usepackage{amsmath} 
\usepackage{xcolor}
\usepackage{multirow} 
\usepackage{algorithm} 
\usepackage{algorithmic} 
\DeclareMathOperator{\crf}{C}
\DeclareMathOperator{\crfoper}{CRF}
\DeclareMathOperator{\mae}{MAE}

\begin{document}
%
\title{Weakly Supervised Salient Object Detection Using 
Image Labels}

\author{Guanbin Li$^1$, Yuan Xie$^1$, Liang Lin$^{1,2}$\thanks{Corresponding author is Liang Lin (Email: linliang@ieee.org). This work was supported in part by the National Natural Science Foundation of China under Grant 61702565, in part by the Special Program of the NSFC-Guangdong Joint Fund for Applied Research on Super Computation (the second phase), in part by Guangdong Natural Science Foundation Project for Research Teams under Grant 2017A030312006. This work was also sponsored by CCF-Tencent Open Research Fund.}\\
$^1$School of Data and Computer Science, Sun Yat-sen University, Guangzhou, China\\
$^2$SenseTime Group Limited\\
}

\maketitle
\begin{abstract}
  Deep learning based salient object detection has recently achieved great success with its performance greatly outperforms any other unsupervised methods. However, annotating per-pixel saliency masks is a tedious and inefficient procedure. In this paper, we note that superior salient object detection can be obtained by iteratively mining and correcting the labeling ambiguity on saliency maps from traditional unsupervised methods. We propose to use the combination of a coarse salient object activation map from the classification network and saliency maps generated from unsupervised methods as pixel-level annotation, and develop a simple yet very effective algorithm to train fully convolutional networks for salient object detection supervised by these noisy annotations. Our algorithm is based on alternately exploiting a graphical model and training a fully convolutional network for model updating. The graphical model corrects the internal labeling ambiguity through spatial consistency and structure preserving while the fully convolutional network helps to correct the cross-image semantic ambiguity and simultaneously update the coarse activation map for next iteration. Experimental results demonstrate that our proposed method greatly outperforms all state-of-the-art unsupervised saliency detection methods and can be comparable to the current best strongly-supervised methods training with thousands of pixel-level saliency map annotations on all public benchmarks. 
\end{abstract}

\section{Introduction}
Salient object detection is designed to accurately detect distinctive regions in an image that attract human attention. Recently, this topic has attracted widespread interest in the research community of computer vision and cognitive science as it can be applied to benefit a wide range of artificial intelligence and vision applications, such as robot intelligent control~\cite{shon2005probabilistic}, content-aware image editing~\cite{avidan2007seam}, visual tracking~\cite{mahadevan2009saliency} and video summarization~\cite{ma2002user}. 

Recently, the deployment of deep convolutional neural networks has resulted in significant progress in salient object detection~\cite{li2016visual,LiYu16,liu2016dhsnet,wang2016saliency}. The performance of these CNN based methods, however, comes at the cost of requiring pixel-wise annotations to generate training data. For salient object detection, it is painstaking to annotate mask-level label and takes several minutes for an  experienced annotator to label one image. Moreover, as the definition of an object being salient is very subjective, there often exists multiple diverse annotations for a same image between different annotators. To ensure the quality of training data sets, these images with ambiguous annotations should be removed, which makes the labeling task more     laborious and time-consuming. This time-consuming task is bound to limit the total amount of pixel-wise training samples and thus become the bottleneck of further development of fully-supervised learning based methods.

\begin{figure}[t]
\begin{center}
   \includegraphics[width=0.47\textwidth]{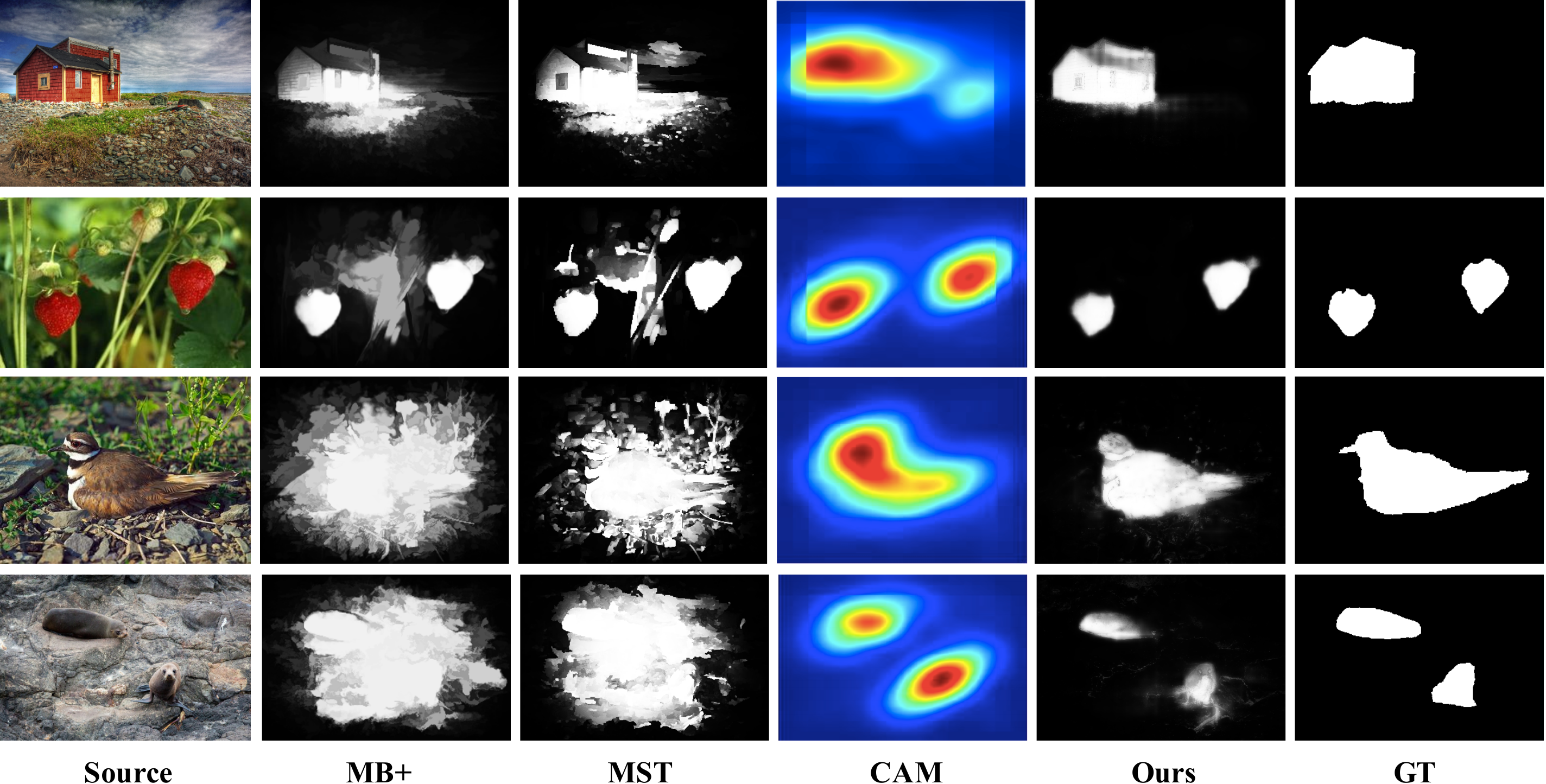}
\end{center}
   \caption{Two kinds of defects in state-of-the-art unsupervised salient object detection methods vs. the results of our proposed weakly supervised optimization framework.}
\label{fig:sample_smap}
\end{figure} 

As a low level vision problem, there exists an ocean of unsupervised salient object detection methods~\cite{wei2012geodesic,cheng2015global,tu2016real,zhang2015minimum,yang2013saliency}. These methods are usually based on low-level features such as color, gradient or contrast and some saliency priors, such as the center prior~\cite{liu2011learning} and the background prior~\cite{wei2012geodesic}. As it is impractical to define a set of universal rules for how an object being salient, each of these bottom-up methods works well for some images, but none of them can handle all the images. 
By observing the failure cases, We found that most of the saliency detection error lies in the lack of spatial correlation inference and image semantic contrast detection. As shown in Figure~\ref{fig:sample_smap}, the two unsupervised saliency detection methods are only able to detect part of the salient objects in the first two cases as they fail to take into account the spatial consistency~(e.g. encouraging
nearby pixels with similar colors to take similar saliency
scores), while for the last two cases, the two methods completely fail to detect the salient objects as these objects are of very low contrast in terms of low-level features~(they are salient in high semantic contrast). These two kinds of failure cases are hard to be found in fully-supervised methods.

During the training and testing of several fully-supervised saliency detection methods~\cite{LiYu16,liu2016dhsnet,wang2016saliency}, we found that a well-trained deep-convolution network without over-fitting can even correct some user annotation error exists in the training samples. We conjecture that a large amount of model parameters contained in deep neural network can be trained to discover the universal rules implied in large scale training samples and thus can help to detect the ambiguity in the annotation mask~(noisy annotation) and figure out a ``correct'' one which being in line with the hidden rules. Moreover, recent work has shown that CNNs being trained on image-level {}labels for classification have remarkable ability to localize the most discriminative region of an image~\cite{zhou2016learning}.

Inspired by these observations, in this paper, we address the weakly supervised salient object detection task using only image-level labels, which in most cases specify salient objects within the image. We develop an algorithm that exploits saliency maps generated from any unsupervised method as noisy annotations to train convolutional networks for better saliency maps. Specifically, we first propose a conditional random field based graphical model to correct the internal label ambiguity by enhancing the spatial coherence and salient object localization. Meanwhile, a multi-task fully convolutional ResNet~\cite{he2015deep} is learned, which is supervised by the iteratively corrected pixel-level annotations as well as image labels~(indicating significant object class within an image), and in turn provides two probability maps to generate an updated unary potential for the graphical model. The first probability map is called Class Activation Map~(CAM) and it highlights the discriminative object parts detected by the image classification-trained CNN while the second one being a more accurate saliency map trained from pixel-wise annotation. Though CAM itself is a relatively coarse pixel-level probability map, it shows very accurate salient object localization ability and thus can be used as a guide to generate more precise pixel-wise annotation for a second round training. The proposed method is optimized alternately until a stopping criteria appears. In our experiment, we find that although CAM is trained using images from a fix number of image classes, it generalizes well to images of unknown categories, resulting in an intensely accurate salient object positioning for generic salient objects. The proposed optimization framework also theoretically applies to all unsupervised salient object detection methods and is able to generate more accurate saliency map very efficiently in fewer than one second per image no matter how time-consuming the original model.

In summary, this paper has the following contributions:
\begin{itemize}
\item We introduce a generic alternate optimization framework to fill the performance gap between supervised and unsupervised salient object detection methods without resorting to laborious pixel labeling.
\item We propose a conditional random field based graphical model to cleanse the noisy pixel-wise annotation by enhancing the spatial coherence as well as salient object localization. 
\item We also design a multi-task fully convolutional ResNet-101 to both generate a coarse class activation map~(CAM) and a pixel-wise saliency probability map, the cooperation of which can help to detect and correct the cross-image annotation ambiguity, generating more accurate saliency annotation for iterative training. 
\end{itemize}

\begin{figure*}[ht]
\begin{center}
   \includegraphics[width=0.80\textwidth]{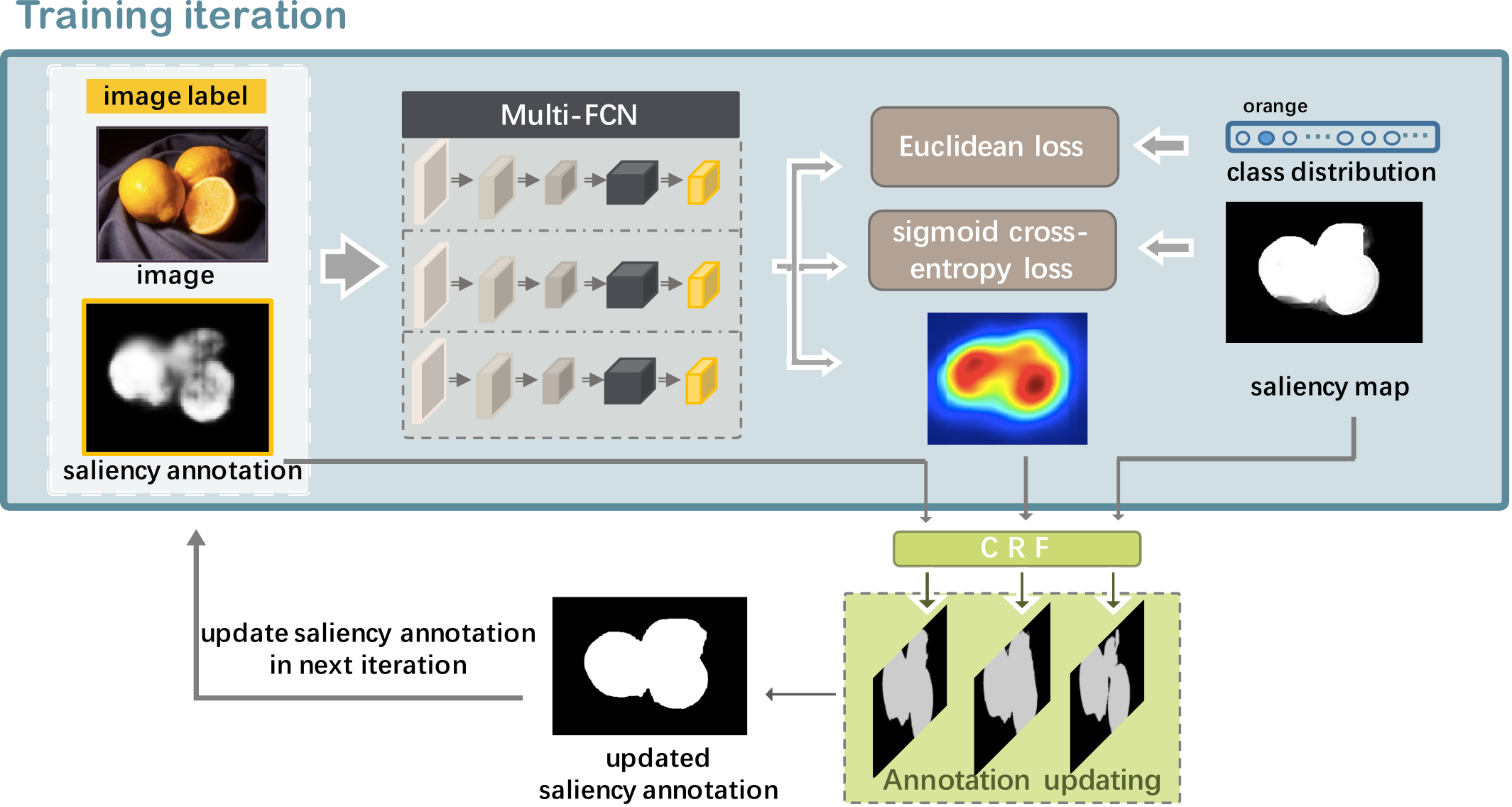}
\end{center}
   \caption{Overall framework for alternate saliency map optimization.}
\label{fig:architecture}
\end{figure*}

\section{Alternate Saliency Map Optimization}\label{sec:alternate_optimization}
As shown in Figure~\ref{fig:architecture}, our proposed saliency map optimization framework consists of two components, a multi-task fully convolutional network~(Multi-FCN) and a graphical model based on conditional random fields~(CRF). Given the Microsoft COCO dataset~\cite{lin2014microsoft} with multiple image labels corresponding to each image, we initially utilize a state-of-the-art unsupervised salient object detection method, i.e.~minimum barrier salient object detection~(MB$+$), to generate the saliency maps of all training images. The produced saliency maps as well as their corresponding image labels are employed to train the Multi-FCN, which simultaneously learns to predict a pixel-wise saliency map and an image class distribution. When training converged, a class activation mapping technique~\cite{zhou2016learning} is applied to the Multi-FCN to generate a serious of class activation maps~(CAMs). Then the initial saliency map, the predicted saliency map from Multi-FCN as well as the average map of the top three CAMs~(CAM prediction corresponding to top $3$ classes) are employed to the CRF model to get the corresponding maps with better spatial coherence and contour localization. We further propose an annotation updating scheme to construct new saliency map annotations from these three maps with CRF for a second iteration of Multi-FCN training. Finally, to generalize the model for saliency detection of unknown image labels, we further finetune the saliency map prediction stream of the Multi-FCN guided by generated CAM using salient object detection datasets~(e.g. MSRA-B and HKUIS) without annotations.
\subsection{Multi-Task Fully Convolutional Network}
In the multi-task fully convolutional stream, we aim to design an end-to-end convolutional network that can be viewed as a combination of the image classification task and the pixel-wise saliency prediction task. To conceive such an end-to-end architecture, we have the following considerations. First, the network should be able to correct the noisy initial saliency annotations as well as possible by mining the semantic ambiguity between the images. Second, the network should be able to be end to end trainable to output a saliency map with appropriate resolution. Last but not the least, it should also be able to detect visual contrast at different scales.

We choose ResNet-101~\cite{he2015deep} as our pre-trained network and modify it to meet our requirements. We first refer to~\cite{chen2014semantic} and  re-purpose it into a dense image saliency prediction network by replacing its 1000-way linear classification
layer with a linear convolutional layer with a $1\times 1$ kernel and two output channels. The feature maps after the final convolutional layer is only $1/32$ of that of the original input image because the original
ResNet-101 consists of one pooling layer and 4 convolutional
layers, each of which has stride 2. We call these five layers
“down-sampling layers”. As described in~\cite{he2015deep}, the 101 layers in ResNet-101 can be divided into five groups. Feature maps computed by different layers in each group share the same resolution. To make the final saliency map denser, we skip subsampling in the last two down-sampling layers by setting their stride to 1, and increase the dilation rate of subsequent convolutional kernels using the dilation algorithm to enlarge their receptive fields as~\cite{chen2014semantic}. Therefore, all the features maps in the last three groups have the same resolution, $1/8$ original resolution after network transformation.

\begin{figure}[ht]
\centerline{
   \includegraphics[width=0.50\textwidth]{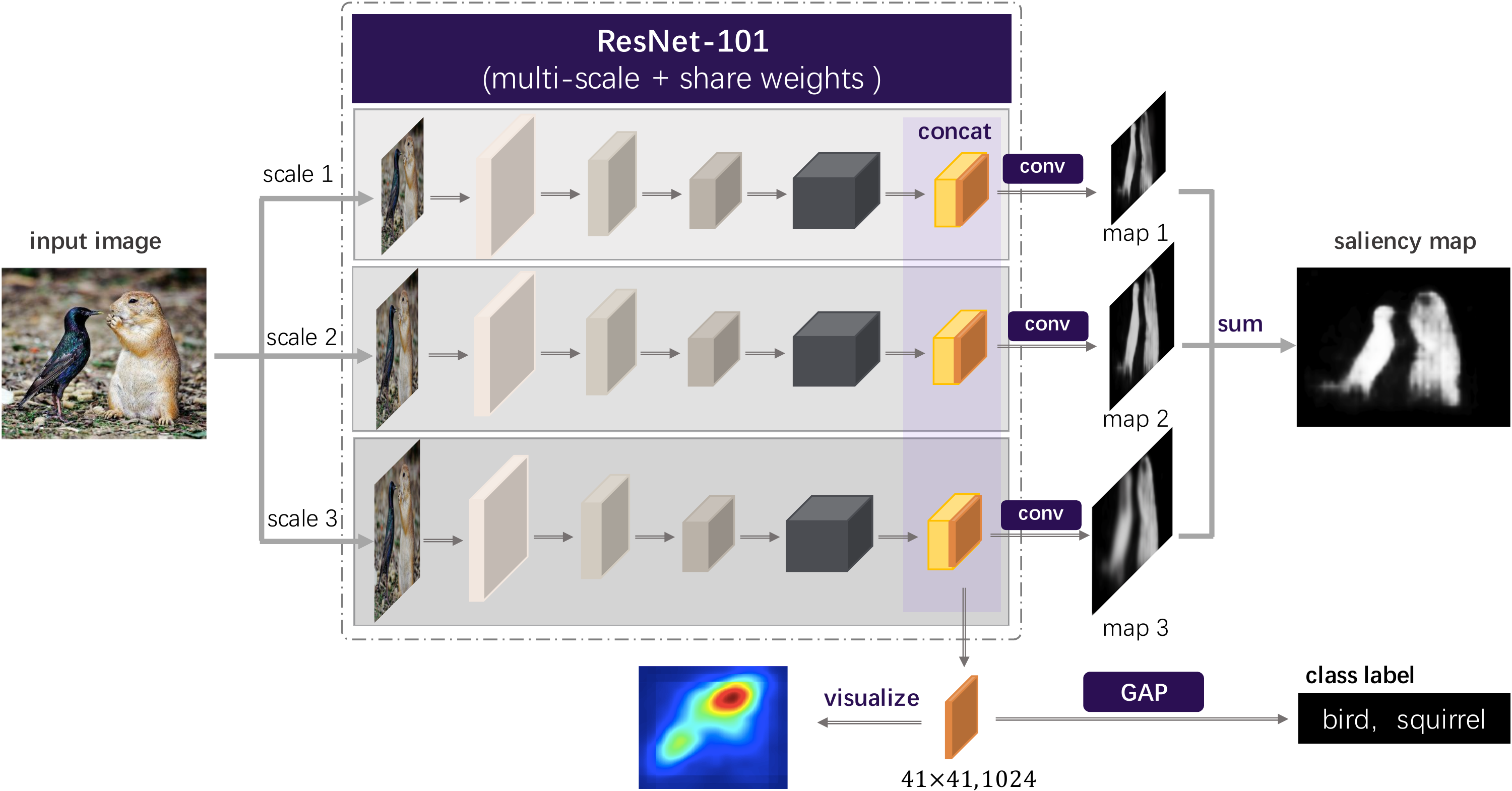}
}
   \caption{The architecture of our multi-task fully convolutional network~(Multi-FCN).\vspace{-2mm}}
\label{fig:multi-fcn}
\end{figure}

As it has been widely verified that feeding multiple scales
of an input image to networks with shared parameters
are rewarding for accurately localizing objects of different scales~\cite{chen2015attention,lin2015efficient}, we replicate the fully convolutional ResNet-101 network three times, each responsible for one input scale $s$~($s\in \{0.5, 0.75, 1\}$). Each scale $s$ of the input image is fed to one of the three replicated ResNet-101, and outputs a two-channel probability map in the resolution of scale $s$, denoted as $M_{c}^{s}$, where $c \in \{0,1\}$ denotes the two classes for saliency detection. The three probability maps are resized to the same resolution as the raw input image using bilinear interpolation, summed up and fed to a sigmoid layer to produce the final probability map. The network framework is shown in Figure~\ref{fig:multi-fcn}.

For image classification task, as we desire to perform object localization from the classification model, we refer to~\cite{zhou2016learning} and integrate a global average pooling layer for generating class activation maps. Specifically, as shown in Figure~\ref{fig:multi-fcn}, we rescale the three output feature maps of the last original convolutional layer in ResNet-101~(corresponds to three input scale) to the same size~($1/8$ original resolution) and concatenate to form feature maps for classification. We further perform global average pooling on the concatenated convolutional feature maps and use those as features for a fully-connected layer
which produces the desired classes distribution output. Let $f_k(x,y)$ represent the activation of channel $k$ in the concatenated feature map at spatial location $(x,y)$. Define $M_c$ as the class activation map for class $c$, where each spatial element can be calculated as follows~\cite{zhou2016learning}:
\begin{equation}
  M_{c}\left(x,y\right) = \sum_{k}w_k^cf_k\left(x,y\right).
\end{equation}
$w_k^c$ is the weight corresponding to class $c$ for unit $k$~(after global average pooling, each channel of the concatenated feature map becomes a unit activation value).

\subsection{Graphical Model for Saliency Map Refinement}
By observing the saliency maps generated by state-of-the-art unsupervised methods, we find that for images with low contrast and complex background, the salient object can hardly be completely detected, with common defects exist in  spatial consistency and contour preserving. We call these defects internal labeling ambiguity in noisy saliency annotations. Fully connect CRF model has been widely used in semantic segmentation~\cite{krahenbuhl2012efficient,chen2014semantic} to both refine the segmentation result and better capture the object boundaries. It has also been used as a post-processing step in~\cite{LiYu16,li2016visual} for saliency map refinement. In this paper, we refer to~\cite{LiYu16} and utilize the same formulation and solver of the two classes fully connected CRF model to correct the internal labeling ambiguity. The output of the CRF operation is a probability map, the value of which denotes the probability of each pixel being salient. We convert it into a binary label by thresholding when being used as training sample annotations.

\subsection{Saliency Annotations Updating Scheme}\label{sec:saliency_update}
We denote the original input image as $I$ and the corresponding saliency map of the specific unsupervised method as $S_{anno}$. After convergence of the first complete training of Multi-FCN, we apply the trained model to generate saliency maps as well as the average map of the top $3$ class activation maps for all training images. We denote the predicted saliency map as $S_{predict}$ and the average class activation map as $S_{cam}$. Furthermore, we also perform fully connected CRF operation to the initial saliency maps produced by a specific unsupervised method, the predicted saliency map from Multi-FCN as well as the average class activation map. The resulting saliency maps are denoted as $\crf_{anno}$, $\crf_{predict}$ and $\crf_{cam}$ respectively. Base on this, we update the training samples as well as their corresponding saliency annotations for the next iteration according to Algorithm~\ref{alg_saliency_anno_update}. $\crfoper\left(\right)$ denotes the CRF operation while $S_{update}$ refers to the updated saliency map annotation, which is further used as the saliency groundtruth for the next iterative training. $\mae\left(\right)$ is defined as the average pixelwise absolute difference between two saliency maps~(i.e. $S_1$ and $S_2$), which is calculated as follows:
\begin{equation}
\mae\left(S_1, S_2\right) = \frac{1}{W\times H}\sum_{x=1}^{W}\sum_{y=1}^{H}|S_1(x,y) - S_2(x,y)|.
\label{equa:mae}
\end{equation}
where $W$ and $H$ being the width and height of the saliency map. 


\begin{algorithm} 
\caption{Saliency Annotations Updating}  
\label{alg_saliency_anno_update}  
\begin{algorithmic}[1]  
\REQUIRE Current saliency map annotation $S_{anno}$, the predicted saliency map $S_{predict}$, CRF output of current saliency map annotation $\crf_{anno}$, CRF output of the predicted saliency map $\crf_{predict}$ and CRF output of the class activation map $\crf_{cam}$ 
\ENSURE The updated saliency map annotation $S_{update}$. 
\IF{$\mae\left(\crf_{anno}, \crf_{predict}\right) \leq \alpha$ } 
\STATE $S_{update} = \crfoper\left(\frac{S_{anno} + S_{predict}}{2}\right)$  
\ELSIF{$\mae\left(\crf_{anno}, \crf_{cam}\right) > \beta$ and    $\mae\left(\crf_{predict}, \crf_{cam}\right) > \beta$ } 
\STATE Discard the training sample in next iteration
\ELSIF {$\mae\left(\crf_{anno}, \crf_{cam}\right) \leq \mae\left(\crf_{predict}, \crf_{cam}\right)$} 
\STATE  $S_{update} = \crf_{anno}$
\ELSE 
\STATE $S_{update} = \crf_{predict}$
\ENDIF

\end{algorithmic} 
\end{algorithm}

\subsection{Multi-FCN Training with Weak Labels}\label{sec:multi-training}
The training of the weakly supervised saliency map optimization framework is composed of two stages, both of which are based on an alternative training scheme. In the first stage, we train the model using Microsoft COCO dataset~\cite{lin2014microsoft} with multiple image labels per image. Firstly, we choose a state-of-the-art unsupervised salient object detection model and apply it to produce an initial saliency map for each image in the training. Then we simply put the saliency maps as initial annotation and train the Multi-FCN for a pixel-wise saliency prediction as well as the classification model for better class activation map generation. While training, we validate on the validation set of the COCO dataset also with generated pixel-wise noisy annotations being the groundtruth. Also note that in order to speed up the training, we initialize the Multi-FCN with a pre-trained model over the ImageNet dataset~\cite{deng2009imagenet} instead of training from scratch. After training convergence, we choose the model with lowest validation error as the final model for this iteration, and apply it to generate saliency maps as well as the average map of top $3$ class activation maps for all training images. Secondly, we apply saliency annotations updating scheme according to Section~\ref{sec:saliency_update} to create updated training tuples~(images, saliency annotation and image label) for a second round of training. We alternately train the model until a stopping criteria appears. After each training round, we evaluate the mean MAE between each pair of saliency annotation~(Pseudo Groundtruth) and the predicted saliency map, and the stopping criteria is defined to be the mean MAE gets lower than a specific threshold or the total number of training rounds reaches $5$.~(Noted that as being a weakly supervised method, we do not use true annotations to determine the merits of the model).

Finally, in order to generalize the model for generic salient object detection with unknown image labels, we further finetune the saliency map prediction stream of the Multi-FCN guided by offline CAMs using salient object detection datasets (e.g. the training images of MSRA-B and HKU-IS) without annotations, until the stopping criteria appears. Here, we calculate the mean of the top $5$ CAMs as the guided CAM, and we discover that although CAM is trained with specific image classification labels, its predicted CAMs of the most similar categories in the category set can still highlight the most discriminative regions in the image and thus still works well as an auxiliary guidance for generic salient object detection. The loss function for updating Multi-FCN for pixel-wise saliency prediction is defined as the sigmoid cross entropy between the generated ground truth~($G$) and the predicted saliency map~($S$):

\begin{equation}
\begin{aligned}
L=&-\beta_i\sum_{i=1}^{|I|}G_i\log P\left ( S_i=1|I_i,W \right ) \\& -  \left ( 1-\beta_i \right )\sum_{i=1}^{|I|}\left ( 1-G_i \right )\log P\left ( S_i=0|I_i,W \right ),
\end{aligned}
\end{equation}

where $W$ denotes the collection of corresponding network parameters in the Multi-FCN, $\beta_i$ is a weight balancing the number of salient pixels and unsalient ones, and $|I|$, $|I|\_$ and $|I|_+$ denote the total number of pixels, unsalient pixels and salient pixels in image $I$, respectively. Then $\beta_i=\frac{|I|\_}{|I|}$ and $1-\beta_i=\frac{|I|_+}{|I|}$. When training for multi-label object classification, we simply employ the Euclidean loss as the objective function and only update the parameters of the fully connected inference layer with parameters of the main ResNet-101 being unchanged.

\begin{figure*}[ht]
    \centerline{
    \includegraphics[width = 0.32\textwidth]{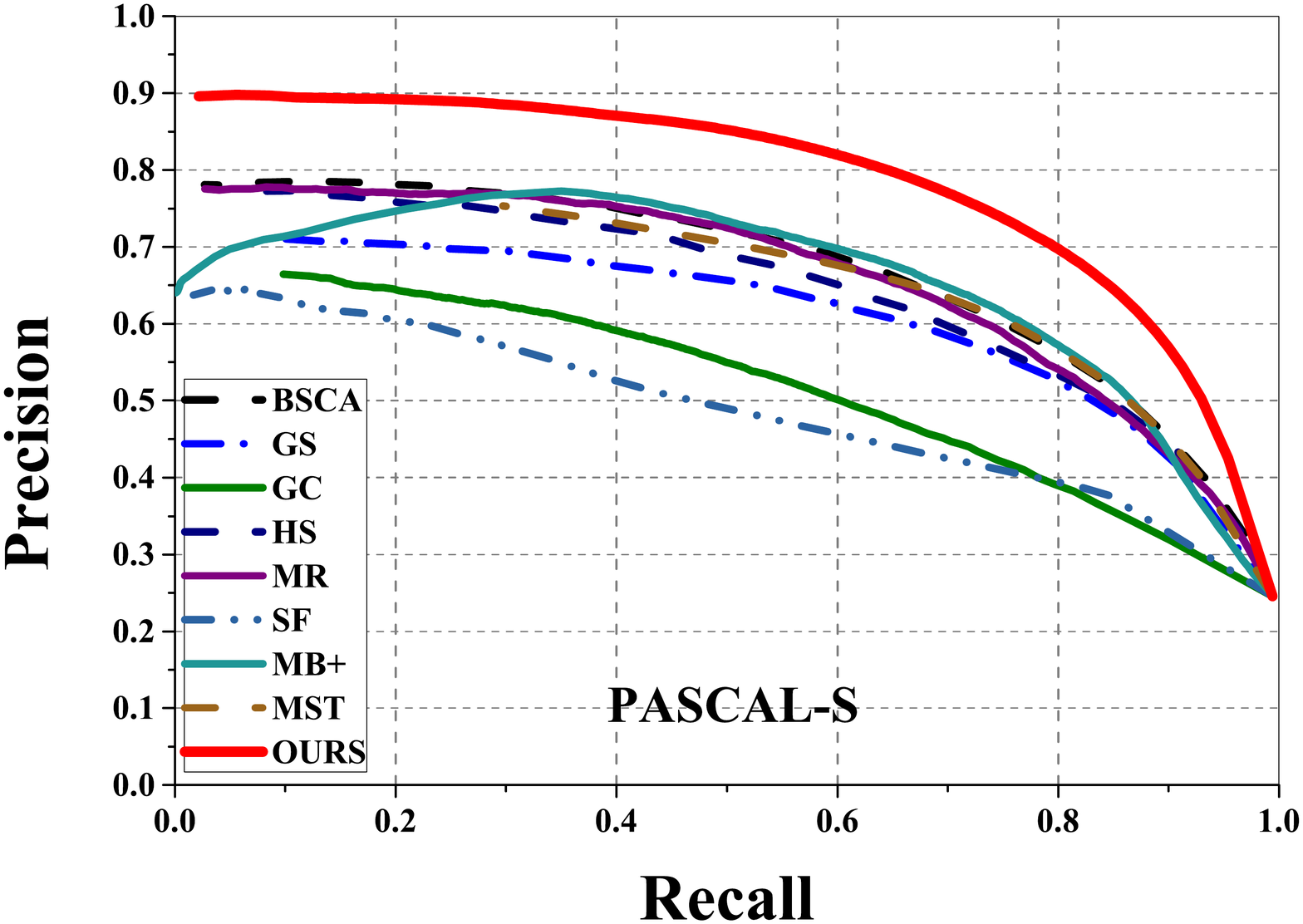}\hfill
    \includegraphics[width = 0.32\textwidth]{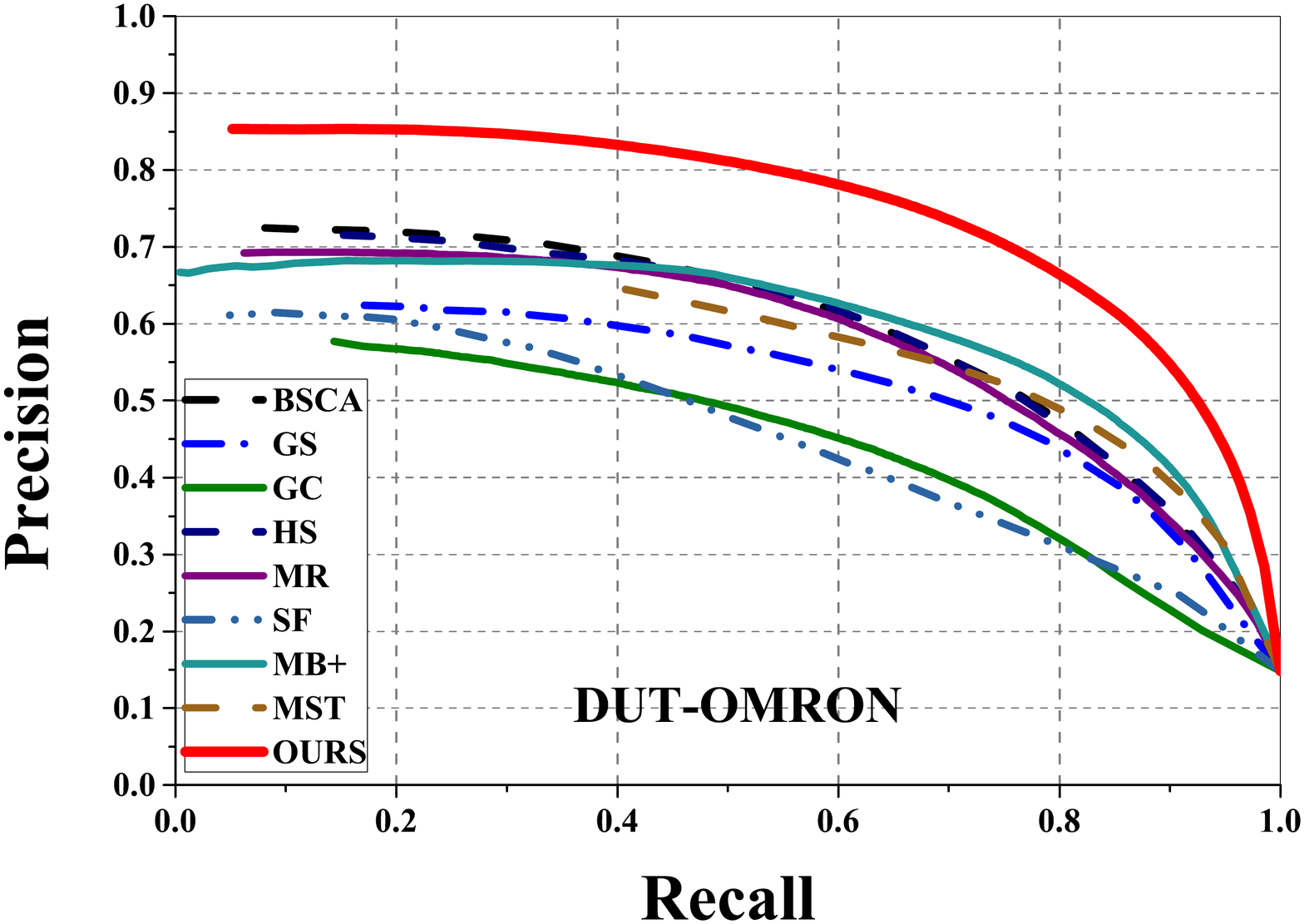}\hfill
    \includegraphics[width = 0.32\textwidth]{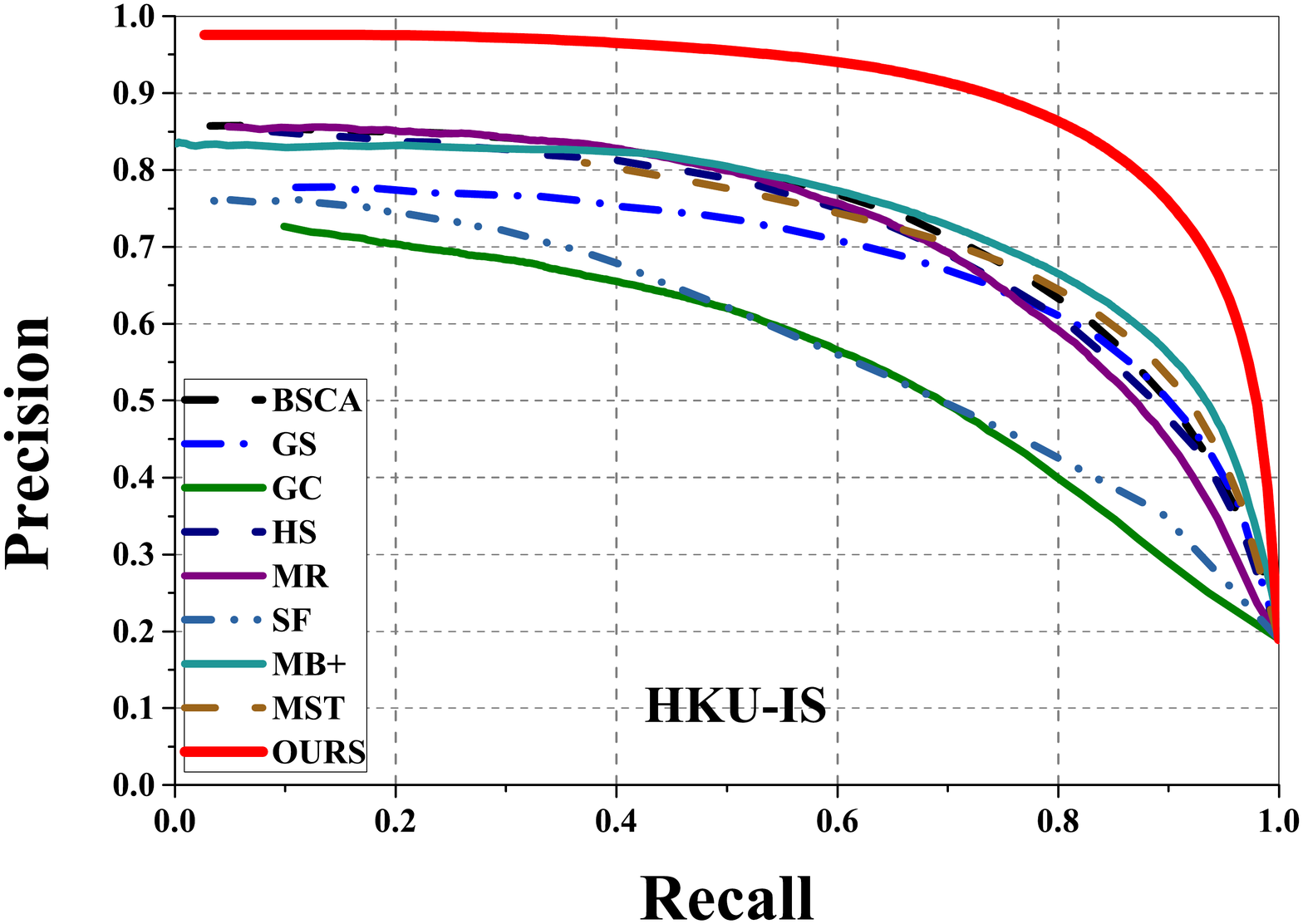}
  }
  \caption{Comparison of precision-recall curves among 9 salient region detection methods on 3 datasets. Our proposed ASMO consistently outperforms other methods across all the testing datasets. \vspace{-1mm}
  }
  \label{fig:comps_pr}
\end{figure*}

\section{Experimental Results}
\subsection{Implementation}
Our proposed Multi-FCN has been implemented on the
public DeepLab code base~\cite{chen2014semantic}. A GTX Titan X GPU is used for both training and testing. As described in Section~\ref{sec:multi-training}, the Multi-FCN involves two stages training. In the first stage, we train on Microsoft COCO object detection dataset for multi-label recognition, which comprises a training set of 82,783 images, and
a validation set of 40,504 images. The dataset covers 80
common object categories, with about 3.5 object labels per
image. In the second stage, we combine the training images of
both the MSRA-B dataset~(2500 images)~\cite{liu2011learning} and the HKU-IS dataset~(2500 images)~\cite{li2016visual} as our training set (5000 images), with all original saliency annotations removed. The validation sets without annotations in the aforementioned two datasets are also combined as our validation set (1000 images). During training, the mini-batch size is set to 2 and we choose to update the loss every 5 iterations. We set the momentum parameter to 0.9 and the weight decay to 0.0005 for both subtasks. The total number of iteration is set to 8K during each training round. During saliency annotation updating, the thresholds $\alpha$ and $\beta$ are set to 15 and 40 respectively. The mean MAE of the training stop criteria is set to 0.05 in our experiment. 

\subsection{Datasets}
We conducted evaluations on six public salient object benchmark datasets:~MSRA-B~\cite{liu2011learning}, PASCAL-S~\cite{li2014secrets}, DUT-OMRON\cite{yang2013saliency}, HKU-IS~\cite{li2016visual}, ECSSD~\cite{yan2013hierarchical} and SOD~\cite{martin2001database}. Though we do not use any user annotations in training, we get to know the training and validation sets of the MSRA-B and HKU-IS datasets in advance. Therefore, for the sake of fairness, we evaluate our model on the testing sets of these two datasets and on the combined training and testing sets of other datasets.

\subsection{Evaluation Metrics}
We adopt precision-recall curves~(PR), maximum F-measure and mean absolute error~(MAE) as our performance measures. The continuous saliency map is binarized using different thresholds varying from 0 to 1. At each threshold value, a pair of precision and recall value can be obtained by comparing the binarized saliency map against the binary groundtruth. The PR curve of a dataset is obtained from all pairs of average precision and recall over saliency maps of all images in the dataset. The F-measure is defined as
  $F_{\beta} = \frac{(1+\beta^2)\cdot Precision \cdot Recall}{\beta^2\cdot Precision + Recall},$
where $\beta^2$ is set to 0.3. We report the maximum F-measure computed from all precision-recall pairs. 
MAE is defined as the average pixelwise absolute difference between the binary ground truth and the saliency map~\cite{perazzi2012saliency} as described in Equation~\ref{equa:mae}.

\begin{table*}[]
\centering
\resizebox{0.92\textwidth}{!}
{

\begin{tabular}{|c|c|c|c|c|c|c|c|c|c|c|c|}
\hline
Data Set                    & Metric & GS    & SF                                    & HS    & MR                                    & GC    & BSCA                                  & MB+                                   & MST                                   & ASMO                                  & ASMO+                                 \\ \hline
                            & maxF   & 0.777 & 0.700                                 & 0.813 & {\color[HTML]{32CB00} \textbf{0.824}} & 0.719 & 0.830                                 & 0.822                                 & 0.809                                 & {\color[HTML]{3531FF} \textbf{0.890}} & {\color[HTML]{FE0000} \textbf{0.896}} \\ \cline{2-12} 
\multirow{-2}{*}{MSRA-B}    & MAE    & 0.144 & 0.166                                 & 0.161 & 0.127                                 & 0.159 & 0.130                                 & 0.133                                 & {\color[HTML]{32CB00} \textbf{0.098}} & {\color[HTML]{FE0000} \textbf{0.067}} & {\color[HTML]{3531FF} \textbf{0.068}} \\ \hline
                            & maxF   & 0.661 & 0.548                                 & 0.727 & 0.736                                 & 0.597 & {\color[HTML]{32CB00} \textbf{0.758}} & 0.736                                 & 0.724                                 & {\color[HTML]{3531FF} \textbf{0.837}} & {\color[HTML]{FE0000} \textbf{0.845}} \\ \cline{2-12} 
\multirow{-2}{*}{ECSSD}     & MAE    & 0.206 & 0.219                                 & 0.228 & 0.189                                 & 0.233 & 0.183                                 & 0.193                                 & {\color[HTML]{32CB00} \textbf{0.155}} & {\color[HTML]{FE0000} \textbf{0.110}} & {\color[HTML]{3531FF} \textbf{0.112}} \\ \hline
                            & maxF   & 0.682 & 0.590                                 & 0.710 & 0.714                                 & 0.588 & 0.723                                 & {\color[HTML]{32CB00} \textbf{0.727}} & 0.707                                 & {\color[HTML]{3531FF} \textbf{0.846}} & {\color[HTML]{FE0000} \textbf{0.855}} \\ \cline{2-12} 
\multirow{-2}{*}{HKU-IS}    & MAE    & 0.166 & 0.173                                 & 0.213 & 0.174                                 & 0.211 & 0.174                                 & 0.180                                 & {\color[HTML]{32CB00} \textbf{0.139}} & {\color[HTML]{FE0000} \textbf{0.086}} & {\color[HTML]{3531FF} \textbf{0.088}} \\ \hline
                            & maxF   & 0.556 & 0.495                                 & 0.616 & 0.610                                 & 0.495 & 0.617                                 & {\color[HTML]{32CB00} \textbf{0.621}} & 0.588                                 & {\color[HTML]{3531FF} \textbf{0.722}} & {\color[HTML]{FE0000} \textbf{0.732}} \\ \cline{2-12} 
\multirow{-2}{*}{DUT-OMRON} & MAE    & 0.173 & {\color[HTML]{32CB00} \textbf{0.147}} & 0.227 & 0.187                                 & 0.218 & 0.191                                 & 0.193                                 & 0.161                                 & {\color[HTML]{3531FF} \textbf{0.101}} & {\color[HTML]{FE0000} \textbf{0.100}} \\ \hline
                            & maxF   & 0.620 & 0.493                                 & 0.641 & 0.661                                 & 0.539 & 0.666                                 & {\color[HTML]{32CB00} \textbf{0.673}} & 0.657                                 & {\color[HTML]{3531FF} \textbf{0.752}} & {\color[HTML]{FE0000} \textbf{0.758}} \\ \cline{2-12} 
\multirow{-2}{*}{PASCAL-S}  & MAE    & 0.223 & 0.240                                 & 0.264 & 0.223                                 & 0.266 & 0.224                                 & 0.228                                 & {\color[HTML]{32CB00} \textbf{0.194}} & {\color[HTML]{FE0000} \textbf{0.152}} & {\color[HTML]{3531FF} \textbf{0.154}} \\ \hline
                            & maxF   & 0.620 & 0.516                                 & 0.646 & 0.636                                 & 0.526 & 0.654                                 & {\color[HTML]{32CB00} \textbf{0.658}} & 0.647                                 & {\color[HTML]{3531FF} \textbf{0.751}} & {\color[HTML]{FE0000} \textbf{0.758}} \\ \cline{2-12} 
\multirow{-2}{*}{SOD}       & MAE    & 0.251 & 0.267                                 & 0.283 & 0.259                                 & 0.284 & 0.251                                 & 0.255                                 & {\color[HTML]{32CB00} \textbf{0.223}} & {\color[HTML]{FE0000} \textbf{0.185}} & {\color[HTML]{3531FF} \textbf{0.187}} \\ \hline
\end{tabular}
}
\caption{Comparison of quantitative results including maximum F-measure (larger is better) and MAE (smaller is better). The best three results on each dataset are shown in \color[HTML]{FE0000}\textbf{red}, \color[HTML]{3166FF}\textbf{blue}\color{black}, and \color[HTML]{32CB00}\textbf{green} \color{black}, respectively.}
\label{tab:comp_quantity}
\end{table*}

\subsection{Comparison with the Unsupervised State-of-the-Art}
Our proposed alternate saliency map optimization framework requires an unsupervised benchmark model as initialization. In this section, we choose the state-of-the-art minimum barrier salient object detection (MB+) method as a baseline and take the optimized model as our final model when compared with other benchmarks. In Section~\ref{sec:method_selection}, we will list more results of our proposed method on other baseline models to demonstrate the effectiveness of our proposed algorithm. 

We compare our method with eight classic or state-of-
the-art unsupervised saliency detection algorithms, including GS~\cite{wei2012geodesic}, SF~\cite{perazzi2012saliency}, HS~\cite{yan2013hierarchical}, MR~\cite{yang2013saliency}, GC~\cite{cheng2015global}, BSCA~\cite{qin2015saliency}, MB+~\cite{zhang2015minimum} and MST~\cite{tu2016real}. For fair comparison, the saliency maps of different methods are provided by authors or obtained from the available implementations.

\begin{figure*}[t]
\centerline{
   \includegraphics[width=0.84\textwidth]{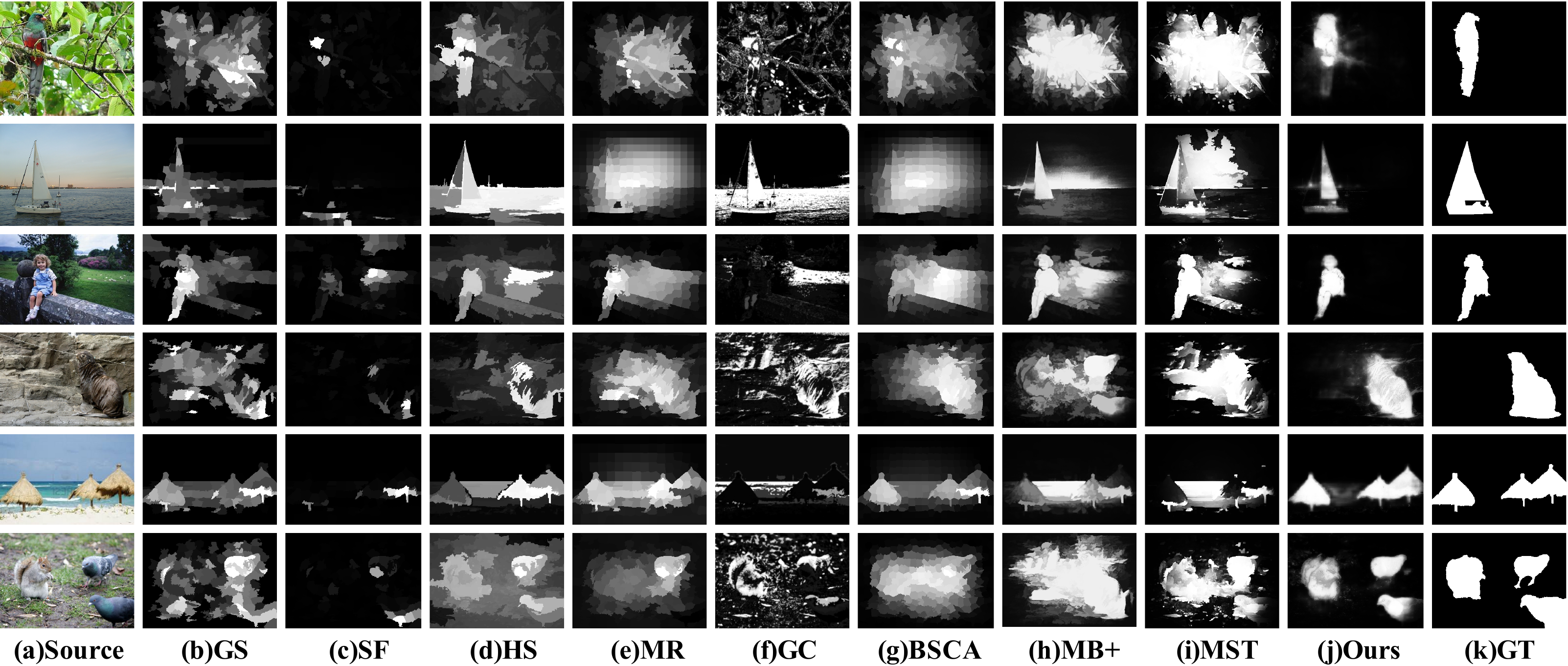}\vspace{-1mm}
}
   \caption{Visual comparison of saliency maps from state-of-the-art methods. The ground truth (GT) is shown in the last column. Our proposed method consistently produces saliency maps closest to the ground truth.\vspace{1mm}
   }
\label{fig:smaps}
\end{figure*}

A visual comparison is given in Fig.~\ref{fig:smaps}. As can be seen, our method generates more accurate saliency maps in various challenging cases, e.g., object in complex background and low contrast between object and background. It is particularly noteworthy that our proposed method employed the saliency maps generated by MB+~\cite{zhang2015minimum} as initial noisy annotations for iterative training, it can learn to mine the ambiguity inside the original noisy labels and the semantic annotation ambiguity across different images, correct them, and eventually produce an optimized results far better than the original ones. As a part of quantitative evaluation, we show a comparison of PR curves in Fig.~\ref{fig:comps_pr}, as shown in the figure, our method significantly outperforms all state-of-the-art unsupervised salient object detection algorithms. 
Moreover, a quantitative comparison of maximum F-measure and MAE is listed in Table.~\ref{tab:comp_quantity}. Our proposed alternate saliency map optimization~(ASMO) improves the maximum F-measure achieved by the best-performing existing algorithm by 8.74\%, 11.48\%, 17.61\%, 17.87\%, 12.63\% and 15.20\% respectively on MSRA-B, ECSSD, HKU-IS, DUT-OMRON,  PASCAL-S and SOD. And at the same time, it lowers the MAE by 31.63\%, 29.03\%, 38.13\%, 31.97\%, 21.65\% and 17.04\% respectively on MSRA-B, ECSSD, HKU-IS, DUT-OMRON, PASCAL-S and SOD. We also evaluate the performance of further applying dense CRF to our proposed method, listed as ASMO+ in the table.


\begin{table*}[ht]

\centering
\caption{Comparison with fully supervised salient object detection methods on HKU-IS, ECSSD and PASCAL-S datasets.}
\label{tab:supervised}
\resizebox{0.85\textwidth}{!}
{

\begin{tabular}{|c|c|c|c|c|c|c|c|c|c|c|}
\hline
Data Set                   & Metric & DRFI  & LEGS                         & MC    & MDF                          & RFCN                         & DHSNet                       & DCL+                         & \multicolumn{1}{l|}{ASMO+} & \begin{tabular}[c]{@{}c@{}}ASMO+\\ (with mask)\end{tabular} \\ \hline
                           & maxF   & 0.776 & {\color[HTML]{333333} 0.770} & 0.798 & 0.861                        & {\color[HTML]{000000} 0.896} & {\color[HTML]{000000} 0.892} & {\color[HTML]{000000} 0.904} & 0.855                      & {\color[HTML]{FE0000} \textbf{0.913}}                       \\ \cline{2-11} 
\multirow{-2}{*}{HKU-IS}   & MAE    & 0.167 & {\color[HTML]{333333} 0.118} & 0.102 & {\color[HTML]{333333} 0.076} & {\color[HTML]{000000} 0.073} & {\color[HTML]{000000} 0.052} & {\color[HTML]{000000} 0.049} & 0.088                      & {\color[HTML]{FE0000} \textbf{0.041}}                       \\ \hline
                           & maxF   & 0.782 & 0.827                        & 0.837 & 0.847                        & {\color[HTML]{000000} 0.899} & {\color[HTML]{000000} 0.907} & {\color[HTML]{000000} 0.901} & 0.845                      & {\color[HTML]{FE0000} \textbf{0.918}}                       \\ \cline{2-11} 
\multirow{-2}{*}{ECSSD}    & MAE    & 0.170 & 0.118                        & 0.100 & 0.106                        & {\color[HTML]{000000} 0.091} & {\color[HTML]{000000} 0.059} & {\color[HTML]{000000} 0.068} & 0.112                      & {\color[HTML]{FE0000} \textbf{0.057}}                       \\ \hline
                           & maxF   & 0.690 & {\color[HTML]{333333} 0.752} & 0.740 & 0.764                        & {\color[HTML]{000000} 0.832} & {\color[HTML]{000000} 0.824} & {\color[HTML]{000000} 0.822} & 0.758                      & {\color[HTML]{FE0000} \textbf{0.847}}                       \\ \cline{2-11} 
\multirow{-2}{*}{PASCAL-S} & MAE    & 0.210 & {\color[HTML]{333333} 0.157} & 0.145 & 0.145                        & {\color[HTML]{000000} 0.118} & {\color[HTML]{000000} 0.094} & {\color[HTML]{000000} 0.108} & 0.154                      & {\color[HTML]{FE0000} \textbf{0.092}}                       \\ \hline
\end{tabular}
}
\end{table*}

\subsection{Ablation Studies}
\subsubsection{Effectiveness of Alternate Saliency Map Optimization} 
Our proposed Multi-FCN based saliency map optimization framework is composed of two components, a multi-task fully convolutional network (Multi-FCN) and a graphical model based on conditional random fields~(CRF). To show the effectiveness of the proposed optimization method, we compare the saliency map $S_1$ generated from the original method, the saliency map $S_2$ from directly employing dense CRF to the original method, the saliency map $S_3$ from training Multi-FCN with generated saliency maps but without CRF or CAM guided, the saliency map $S_4$ from training Multi-FCN with generated saliency maps and CRF guided but without CAM and the saliency map $S_5$ from our full pipeline using DUT-OMRON dataset. As shown in Tab.~\ref{tab:effectiveness}, employing dense CRF operation and iterative training on fully convolutional ResNet-101 with original saliency map as noisy groundtruth can both boost the performance of the original unsupervised method. Our alternately updating scheme can integrate both of these two complementary advantages which further gains 5.22\% improvement on maximum F-measure and 16.6\% decrease on MAE. Moreover, CAM guided saliency annotations updating scheme plays a paramount role in our optimization framework which also greatly improve the saliency map performance.

\begin{table}[]

\centering
\caption{Effectiveness evaluation of different components of alternate saliency map optimization on DUT-OMRON dataset.}
\label{tab:effectiveness}
\resizebox{0.45\textwidth}{!}
{
\begin{tabular}{c|c|c|c|c|c|c}
\hline
Metric & $S_1$    & $S_2$    & $S_3 $   & $S_4$    & $S_5$    & $S_5$+CRF \\ \hline
maxF   & 0.588 & 0.630 & 0.651 & 0.685 & 0.722 & 0.732  \\ \hline
MAE    & 0.161 & 0.178 & 0.151 & 0.126 & 0.101 & 0.100  \\ \hline
\end{tabular}
}
\end{table}

\subsubsection{Sensitivities to Benchmark Method Selection}\label{sec:method_selection}
As described in Section~\ref{sec:alternate_optimization}, our proposed alternate saliency map optimization method is based on an unsupervised benchmark model as initialization. To demonstrate that our proposed method is widely applicable to the optimization of any unsupervised salient object detection method, we apply our optimization method to the other two recently published unsupervised saliency detection methods, including BSCA~\cite{qin2015saliency} and MST~\cite{tu2016real}. Experimental results in Tab.~\ref{tab:benchmark_selection} have shown that although these methods have achieved very good results, the application of our proposed optimization algorithm can still significantly improve their performance. 

\subsubsection{Evaluation on Semi-Supervised Setting}
In this section, we aim to compare our proposed method with the state-of-the-art fully supervised methods. As shown in Tab.~\ref{tab:supervised}, our proposed weakly supervised ASMO with CRF already consistently outperforms 3 fully supervised methods including DRFI~\cite{jiang2013salient}, LEGS~\cite{wang2015deep} and MC~\cite{zhao2015saliency}, and it is comparable to MDF~\cite{li2016visual}. Particularly noteworthy that when we add the groundtruth mask of the training set of MSRA-B dataset to form a semi-supervised setting of our method, it greatly outperforms all state-of-the-art fully supervised methods across all the three testing datasets~(HKU-IS, ECSSD and PASCAL-S). We conjecture that our model considered more semantic information than existing fully-supervised models as we additionally included the Microsoft COCO dataset in our initial training. 

\begin{table}[]

\centering
\caption{Evaluation of different benchmark methods in alternate saliency map optimization on DUT-OMRON dataset.}
\label{tab:benchmark_selection}
\resizebox{0.45\textwidth}{!}
{

\begin{tabular}{c|c|c|c|c|c|c}
\hline
Metric & MB+   & \begin{tabular}[c]{@{}c@{}}ASMO\\ (MB+)\end{tabular} & BSCA  & \begin{tabular}[c]{@{}c@{}}ASMO\\ (BSCA)\end{tabular} & MST   & \begin{tabular}[c]{@{}c@{}}ASMO\\ (MST)\end{tabular} \\ \hline
maxF   & 0.621 & 0.722                                                & 0.617 & 0.685                                                 & 0.588 & 0.691                                                \\ \hline
MAE    & 0.193 & 0.101                                                & 0.191 & 0.121                                                 & 0.161 & 0.126                                                \\ \hline
\end{tabular}}
\end{table}

\section{Conclusions}
In this paper, we have introduced a generic alternate optimization framework to improve the saliency map quality of any unsupervised salient object detection methods by alternately exploiting a graphical model and training a multi-task fully convolutional network for model updating. Experimental results demonstrate that our proposed method greatly outperforms all state-of-the-art unsupervised saliency detection methods and can be comparable to the current best strongly-supervised methods training with thousands of pixel-level saliency map annotations on all public benchmarks.

\bibliographystyle{aaai}
\bibliography{weak_saliency}

\begin{thebibliography}{}

\bibitem[\protect\citeauthoryear{Avidan and Shamir}{2007}]{avidan2007seam}
Avidan, S., and Shamir, A.
\newblock 2007.
\newblock Seam carving for content-aware image resizing.
\newblock {\em ACM Transactions on graphics} 26(3):10.

\bibitem[\protect\citeauthoryear{Chen \bgroup et al\mbox.\egroup
  }{2014}]{chen2014semantic}
Chen, L.-C.; Papandreou, G.; Kokkinos, I.; Murphy, K.; and Yuille, A.~L.
\newblock 2014.
\newblock Semantic image segmentation with deep convolutional nets and fully
  connected crfs.
\newblock {\em arXiv preprint arXiv:1412.7062}.

\bibitem[\protect\citeauthoryear{Chen \bgroup et al\mbox.\egroup
  }{2015}]{chen2015attention}
Chen, L.-C.; Yang, Y.; Wang, J.; Xu, W.; and Yuille, A.~L.
\newblock 2015.
\newblock Attention to scale: Scale-aware semantic image segmentation.
\newblock {\em arXiv preprint arXiv:1511.03339}.

\bibitem[\protect\citeauthoryear{Cheng \bgroup et al\mbox.\egroup
  }{2015}]{cheng2015global}
Cheng, M.-M.; Mitra, N.~J.; Huang, X.; Torr, P.~H.; and Hu, S.-M.
\newblock 2015.
\newblock Global contrast based salient region detection.
\newblock {\em IEEE Transactions on Pattern Analysis and Machine Intelligence}
  37(3):569--582.

\bibitem[\protect\citeauthoryear{Deng \bgroup et al\mbox.\egroup
  }{2009}]{deng2009imagenet}
Deng, J.; Dong, W.; Socher, R.; Li, L.-J.; Li, K.; and Fei-Fei, L.
\newblock 2009.
\newblock Imagenet: A large-scale hierarchical image database.
\newblock In {\em Proceedings of IEEE Computer Vision and Pattern Recognition},
   248--255.

\bibitem[\protect\citeauthoryear{He \bgroup et al\mbox.\egroup
  }{2015}]{he2015deep}
He, K.; Zhang, X.; Ren, S.; and Sun, J.
\newblock 2015.
\newblock Deep residual learning for image recognition.
\newblock {\em arXiv preprint arXiv:1512.03385}.

\bibitem[\protect\citeauthoryear{Jiang \bgroup et al\mbox.\egroup
  }{2013}]{jiang2013salient}
Jiang, H.; Wang, J.; Yuan, Z.; Wu, Y.; Zheng, N.; and Li, S.
\newblock 2013.
\newblock Salient object detection: A discriminative regional feature
  integration approach.
\newblock In {\em Proceedings of IEEE Computer Vision and Pattern Recognition},
   2083--2090.

\bibitem[\protect\citeauthoryear{Kr{\"a}henb{\"u}hl and
  Koltun}{2012}]{krahenbuhl2012efficient}
Kr{\"a}henb{\"u}hl, P., and Koltun, V.
\newblock 2012.
\newblock Efficient inference in fully connected crfs with gaussian edge
  potentials.
\newblock {\em arXiv preprint arXiv:1210.5644}.

\bibitem[\protect\citeauthoryear{Li and Yu}{2016a}]{LiYu16}
Li, G., and Yu, Y.
\newblock 2016a.
\newblock Deep contrast learning for salient object detection.
\newblock In {\em Proceedings of the IEEE Conference on Computer Vision and
  Pattern Recognition},  478--487.

\bibitem[\protect\citeauthoryear{Li and Yu}{2016b}]{li2016visual}
Li, G., and Yu, Y.
\newblock 2016b.
\newblock Visual saliency detection based on multiscale deep cnn features.
\newblock {\em IEEE Transactions on Image Processing} 25(11):5012--5024.

\bibitem[\protect\citeauthoryear{Li \bgroup et al\mbox.\egroup
  }{2014}]{li2014secrets}
Li, Y.; Hou, X.; Koch, C.; Rehg, J.~M.; and Yuille, A.~L.
\newblock 2014.
\newblock The secrets of salient object segmentation.
\newblock In {\em Proceedings of IEEE Computer Vision and Pattern Recognition},
   280--287.

\bibitem[\protect\citeauthoryear{Lin \bgroup et al\mbox.\egroup
  }{2014}]{lin2014microsoft}
Lin, T.-Y.; Maire, M.; Belongie, S.; Hays, J.; Perona, P.; Ramanan, D.;
  Doll{\'a}r, P.; and Zitnick, C.~L.
\newblock 2014.
\newblock Microsoft coco: Common objects in context.
\newblock In {\em Proceedings of the European Conference on Computer Vision},
  740--755.

\bibitem[\protect\citeauthoryear{Lin \bgroup et al\mbox.\egroup
  }{2015}]{lin2015efficient}
Lin, G.; Shen, C.; Reid, I.; et~al.
\newblock 2015.
\newblock Efficient piecewise training of deep structured models for semantic
  segmentation.
\newblock {\em arXiv preprint arXiv:1504.01013}.

\bibitem[\protect\citeauthoryear{Liu and Han}{2016}]{liu2016dhsnet}
Liu, N., and Han, J.
\newblock 2016.
\newblock Dhsnet: Deep hierarchical saliency network for salient object
  detection.
\newblock In {\em Proceedings of the IEEE Conference on Computer Vision and
  Pattern Recognition},  678--686.

\bibitem[\protect\citeauthoryear{Liu \bgroup et al\mbox.\egroup
  }{2011}]{liu2011learning}
Liu, T.; Yuan, Z.; Sun, J.; Wang, J.; Zheng, N.; Tang, X.; and Shum, H.-Y.
\newblock 2011.
\newblock Learning to detect a salient object.
\newblock {\em IEEE Transactions on Pattern Analysis and Machine Intelligence}
  33(2):353--367.

\bibitem[\protect\citeauthoryear{Ma \bgroup et al\mbox.\egroup
  }{2002}]{ma2002user}
Ma, Y.-F.; Lu, L.; Zhang, H.-J.; and Li, M.
\newblock 2002.
\newblock A user attention model for video summarization.
\newblock In {\em Proceedings of the tenth ACM international conference on
  Multimedia},  533--542.

\bibitem[\protect\citeauthoryear{Mahadevan and
  Vasconcelos}{2009}]{mahadevan2009saliency}
Mahadevan, V., and Vasconcelos, N.
\newblock 2009.
\newblock Saliency-based discriminant tracking.
\newblock In {\em Proceedings of the IEEE Conference on Computer Vision and
  Pattern Recognition},  1007--1013.

\bibitem[\protect\citeauthoryear{Martin \bgroup et al\mbox.\egroup
  }{2001}]{martin2001database}
Martin, D.; Fowlkes, C.; Tal, D.; and Malik, J.
\newblock 2001.
\newblock A database of human segmented natural images and its application to
  evaluating segmentation algorithms and measuring ecological statistics.
\newblock In {\em Proceedings of International Conference on Computer Vision},
  volume~2,  416--423.

\bibitem[\protect\citeauthoryear{Perazzi \bgroup et al\mbox.\egroup
  }{2012}]{perazzi2012saliency}
Perazzi, F.; Kr{\"a}henb{\"u}hl, P.; Pritch, Y.; and Hornung, A.
\newblock 2012.
\newblock Saliency filters: Contrast based filtering for salient region
  detection.
\newblock In {\em Proceedings of IEEE Computer Vision and Pattern Recognition},
   733--740.

\bibitem[\protect\citeauthoryear{Qin \bgroup et al\mbox.\egroup
  }{2015}]{qin2015saliency}
Qin, Y.; Lu, H.; Xu, Y.; and Wang, H.
\newblock 2015.
\newblock Saliency detection via cellular automata.
\newblock In {\em Proceedings of IEEE Computer Vision and Pattern Recognition},
   110--119.

\bibitem[\protect\citeauthoryear{Shon \bgroup et al\mbox.\egroup
  }{2005}]{shon2005probabilistic}
Shon, A.~P.; Grimes, D.~B.; Baker, C.~L.; Hoffman, M.~W.; Zhou, S.; and Rao,
  R.~P.
\newblock 2005.
\newblock Probabilistic gaze imitation and saliency learning in a robotic head.
\newblock In {\em Proceedings of the IEEE International Conference on Robotics
  and Automation},  2865--2870.

\bibitem[\protect\citeauthoryear{Tu \bgroup et al\mbox.\egroup
  }{2016}]{tu2016real}
Tu, W.-C.; He, S.; Yang, Q.; and Chien, S.-Y.
\newblock 2016.
\newblock Real-time salient object detection with a minimum spanning tree.
\newblock In {\em Proceedings of IEEE Computer Vision and Pattern Recognition},
   2334--2342.

\bibitem[\protect\citeauthoryear{Wang \bgroup et al\mbox.\egroup
  }{2015}]{wang2015deep}
Wang, L.; Lu, H.; Ruan, X.; and Yang, M.-H.
\newblock 2015.
\newblock Deep networks for saliency detection via local estimation and global
  search.
\newblock In {\em Proceedings of IEEE Computer Vision and Pattern Recognition},
   3183--3192.

\bibitem[\protect\citeauthoryear{Wang \bgroup et al\mbox.\egroup
  }{2016}]{wang2016saliency}
Wang, L.; Wang, L.; Lu, H.; Zhang, P.; and Ruan, X.
\newblock 2016.
\newblock Saliency detection with recurrent fully convolutional networks.
\newblock In {\em Proceedings of the European Conference on Computer Vision},
  825--841.

\bibitem[\protect\citeauthoryear{Wei \bgroup et al\mbox.\egroup
  }{2012}]{wei2012geodesic}
Wei, Y.; Wen, F.; Zhu, W.; and Sun, J.
\newblock 2012.
\newblock Geodesic saliency using background priors.
\newblock In {\em Proceedings of the European Conference on Computer Vision},
  29--42.

\bibitem[\protect\citeauthoryear{Yan \bgroup et al\mbox.\egroup
  }{2013}]{yan2013hierarchical}
Yan, Q.; Xu, L.; Shi, J.; and Jia, J.
\newblock 2013.
\newblock Hierarchical saliency detection.
\newblock In {\em Proceedings of IEEE Computer Vision and Pattern Recognition},
   1155--1162.

\bibitem[\protect\citeauthoryear{Yang \bgroup et al\mbox.\egroup
  }{2013}]{yang2013saliency}
Yang, C.; Zhang, L.; Lu, H.; Ruan, X.; and Yang, M.-H.
\newblock 2013.
\newblock Saliency detection via graph-based manifold ranking.
\newblock In {\em Proceedings of IEEE Computer Vision and Pattern Recognition},
   3166--3173.

\bibitem[\protect\citeauthoryear{Zhang \bgroup et al\mbox.\egroup
  }{2015}]{zhang2015minimum}
Zhang, J.; Sclaroff, S.; Lin, Z.; Shen, X.; Price, B.; and Mech, R.
\newblock 2015.
\newblock Minimum barrier salient object detection at 80 fps.
\newblock In {\em Proceedings of International Conference on Computer Vision},
  1404--1412.

\bibitem[\protect\citeauthoryear{Zhao \bgroup et al\mbox.\egroup
  }{2015}]{zhao2015saliency}
Zhao, R.; Ouyang, W.; Li, H.; and Wang, X.
\newblock 2015.
\newblock Saliency detection by multi-context deep learning.
\newblock In {\em Proceedings of IEEE Computer Vision and Pattern Recognition},
   1265--1274.

\bibitem[\protect\citeauthoryear{Zhou \bgroup et al\mbox.\egroup
  }{2016}]{zhou2016learning}
Zhou, B.; Khosla, A.; Lapedriza, A.; Oliva, A.; and Torralba, A.
\newblock 2016.
\newblock Learning deep features for discriminative localization.
\newblock In {\em Proceedings of IEEE Computer Vision and Pattern Recognition},
   2921--2929.

\end{thebibliography}

\end{document}